\documentclass{article}

\usepackage{arxiv}

\usepackage[utf8]{inputenc} 
\usepackage[T1]{fontenc}    
\usepackage{hyperref}       
\usepackage{url}            
\usepackage{booktabs}       
\usepackage{amsfonts}       
\usepackage{nicefrac}       
\usepackage{microtype}      
\usepackage{lipsum}		
\usepackage{graphicx}
\usepackage{natbib}
\usepackage{doi}

\usepackage[inline]{enumitem}
\usepackage{ragged2e}
\usepackage{xcolor}
\usepackage{colortbl}

\usepackage{tabularx}

\title{RoboCupSoccer Review: The Goalkeeper, a Distinctive Player}

\date{} 					

\author{
    Antoine Dizet \\
    Lab-STICC \\
    ENIB, CNRS UMR 6285, CERVVAL \\
    Plouzane, France\\
    dizet@enib.fr \\
    \And
    Ubbo Visser \\
    University of Miami \\
    Coral Gables FL 33146 \\
    USA \\
    \And
    Cedric Buche \\
    CROSSING \\
    ENIB, CNRS IRL 2010 \\
    Adelaide, Australia\\
}

\renewcommand{\shorttitle}{RoboCupSoccer review}

\begin{document}
\maketitle

\begin{abstract}
This article offers a literature review of goalkeeper robots in the context of the RoboCupSoccer competition. The latter is one of the various league categories hosted by the RoboCup Federation, which fosters AI and Robotics with their landmark challenges. Despite the number of articles on the subject of the goalkeeper, there is a lack of studies offering a comprehensive and up-to-date analysis. We propose  to provide a review of  research related to goalkeepers within the RoboCupSoccer leagues in order to extract possible improvements and scientific issues. The goalkeeper, although being a specific player, has many skills in common with other players. Therefore, this review is divided into three parts: perception, cognition and action, where the perception and action parts are common to all players and the cognition part focuses on goalkeepers. The discussion will open up on the possible improvements of the developments made for these goalkeepers.
\end{abstract}

\keywords{RoboCup \and Soccer \and Goalkeeper \and Review \and Robotic}

\section{Introduction}
 RoboCup is an annual robotics competition divided into different sub-competitions, such as RoboCup@Home which aims to develop personal assistance robots or RoboCupRescue in which researchers aim to develop rescue robots capable of navigating through rough terrain or rubble.
 RoboCupSoccer is divided into different leagues. Rolling Robot Leagues, the RoboCup Small Size League (SSL) focused on multi-agent coordination and control, the RoboCup Middle Size League (MSL) focused on mechatronics design, control and multi-agent cooperation. Humanoid Leagues, the Humanoid League and the Standard Platform League (SPL) where the robot must walk, run and kick the ball while maintaining balance, maintaining this balance therefore implies different average speeds compared to Rolling Robot Leagues (below 10 cm/s in Humanoid Leagues and over 1 m/s in Rolling Robot Leagues). And Simulation Leagues, 2D and 3D, focus on artificial intelligence and team strategy. In the context of a RoboCupSoccer match, the aim is not only to score goals but also to prevent the opposite team from scoring. This is the role of the goalkeeper, a specific player to whom we will pay our attention in this article. Some teams are good at scoring a lot of goals, but their goalkeeper takes even more goals. Other teams, on the contrary, are poorly performing in front of the opposite goal, but take less goals themselves. The latter case is ultimately more favourable. In the end, the goalkeeper is one of the key players.

What are the most relevant research areas and what is the state of the art of goalkeeper-related developments? Our research is based on articles of the past five years of RoboCup proceedings (2015-2019) \citep{chalup2019robocup,holz2018robocup,akiyama2017robocup,behnke2016robocup,almeida2015robocup}. The literature offers fourteen articles related to the goalkeeper with an increase in the number of publications on the subject over the past couple of years. Despite the number of articles on this subject, no study offers a state of the art review.

The goalkeeper is a particular player requiring specific developments. Nevertheless, the goalkeeper has common abilities with other players. As this paper is a review we will start by describe the
\begin{enumerate*}
\item Perception (section \ref{Perception}) which offers the possibility for the all robots to recognise and understand its environment. Then we will discuss about the goalkeeper's \item Cognition (section \ref{Cognition}) including decision making algorithms. Although the developments in terms of cognition are described here in the context of a review on goalkeepers, the methods mentioned can mostly be used on other robots. Once the decision based on the perception is made the robot need to act. This is why the last part of this review will enumerate some of the most important \item Actions (section \ref{Action}) developments such as the navigation for bipedal robots (section \ref{Walking}) which allows the robot to move in this perceived environment, the kick (section \ref{Kicking}) which is a skill both offensively, to score a goal, and defensively, to clear the ball from a dangerous position or to make a pass at a teammate as well as some specific goalkeeping moves (section \ref{Goalkepper motions}).
\end{enumerate*}
In order to have an overview of the different developments made in each of these areas, they are grouped in the tables (cf. Table \ref{table:RoboCupSoccer general developments} and \ref{table:RoboCupSoccer Goalkeeper developments}). These tables are intended to summarize what we will describe in this paper.


\section{Perception}
\label{Perception}
Perception is one of the most important skills. Robot's perception allows to get information on the environment in which it is going to play. The size of this environment (the football field) is not the same in all leagues (SSL: 9mx6m, MSL: 18mx12m, SPL: 10.4mx7.4m). The predominant methods are based on vision, which we will be focusing on in this paper. To do this, algorithms perform image processing on the basis of the video stream from the robot's camera for example. We can divide this vision into two parts: detection and localisation.

Detection analyses the objects present in the robot's field of vision such as other robots, the ball or the opponent's goal (cf. Table \ref{table:RoboCupSoccer general developments}). Although there are many high-performance computer vision algorithms available today (e.g. Mask R-CNN \citep{Kaiming2017Mask}), our use in this case can be limited by the computing capacity of the robot and the video quality of the on-board camera, although this is improving from year to year. In order to overcome the problems due to the on-board camera (low quality and/or positioning) Middle Size League teams can use a Kinect as the hardware platform \citep{luo2017object}. Neural networks and available algorithms have therefore been adapted to these embedded systems. Robot and ball detection for example can be performed via a DNN (Deep Neural Network) having a better performance than Tiny YOLO in terms of speed and accuracy \citep{szemenyei2019robo}. A YOLO-LITE architecture can be also use to reduce computational cost and power consumption \citep{ZhengBai2021faster}. The use of a HOG (Histogram of Oriented Gradient) allows to obtain an average detection accuracy of 88\% for robots in the field of view of a NAO V4 \citep{polceanu2018fast} and of the IGUS humanoid robot \citep{farazi2016real}. The most widely used and most efficient technology in terms of vision remains nevertheless the CNN (Convolutional Neural Network) adapted to the embedded system. Different training strategies for these networks have been proposed. \citep{cruz2017using} reduce the process time to 1ms for a detection rate around 97\% using two detectors, one based on the XNOR-Net and the other on the SqueezeNet. The neural network, used in humanoid league, published by \citep{speck2016ball} determines the x and y coordinates of a ball within the image with an accuracy of 81\% for the x coordinate and 74\% for the y coordinate. It should be noted that the different algorithms mentioned above are found on different robots and therefore have at the same time different computing capacities, so the algorithms will not have the same results on different robots. Nevertheless, regardless of the robot, the published results have a sufficient accuracy to detect the robots and the ball. To go deeper in the robot detection, it is possible to perform a pose estimation of robots \citep{zappel20216d, amini2021real} or a real-time multi-object tracking on stochastic and highly dynamic environments \citep{dias2017real}.

Detection alone does not allow the robot to fully understand the environment. The different objects detected must be positioned in world coordinates. Localisation allows to position the detected objects both in world coordinates and local coordinates. Robots can be positioned thanks to various algorithms. Several solutions are proposed in the literature in order to localise them, a Multiple Model Extended Kalman filter (MM-EKF) based localisation (the filter is applied to the camera output of a RoboCup SPL goalkeeper) \citep{quinlan2009multiple}, a Particle filter based method with a Particle Swarm Optimisation (PSO) \citep{burchardt2010optimizing} for self localisation or a GM-PHD with an error of 20 cm and a calculation time of 0.13 ms \citep{cano2016robust}.This example illustrates well the problem of computing capacity of robots, it is necessary to find the good balance between computing time and performance. An average position error of 20 cm on a surface of 54 m\textsuperscript{2} \citep{robocup2019robocup} for a calculation time of 0.13ms is a good balance. Triangulation was used in RoboCupSoccer MSL (Middle Size League) to position the detected ball in a 3D space with an error of 0.42 m when the ball is in the air and 0.11 m when it is on the ground \citep{kuijpers2016cooperative}. Field boundary detection is mainly used for (self) localisation, \citep{Hasselbring2021Soccer} proposes a CNN based solution with code and data available online. Based on a field detection by Colour segmentation and Histogram analysis, \citep{qian2016adaptive} proposes a method of self-localisation for a NAO V5 (cf. Table \ref{table:RoboCupSoccer general developments}). The use of complex algorithm is significant, however not mandatory as shown by the use of triangulation to locate the ball. To conclude with localisation, based on information on the position of the ball it is possible to perform a prediction on the future position of the ball using a regression model, \citep{mirmohammad2021ball} used a combination of k-NN Regression and Autoregression methods to do it.

A good perception of the environment allows the goalkeeper to have access to enough information to make the right decision at the right time. With this information he is able to evaluate the situation in order to prevent the opposing team from scoring a goal and make the best possible decision. The literature proposes several methods of decision making for goalkeepers (cf. Section \ref{Cognition}).


\section{Cognition}
\label{Cognition}
Once the environment is correctly perceived by the robot, it has the necessary information to make a decision. The literature proposes several multi-robot Systems works based on positioning \citep{neves2015new}, task allocation \citep{dai2019task}, leader election \citep{dias2018multi} and others to set up team strategies. In order to study the cognition of a single agent  robot playing soccer, we will focus on the developments made on the goalkeepers (cf. Table \ref{table:RoboCupSoccer Goalkeeper developments}). Although his specific role differentiates him slightly from his teammates most of the methods we are going to discuss can also be used on non-goalkeeper robots with the difference that a poor decision making on a goalkeeper can allow the opposing team to score a goal and thus win the match.
As a player like the others, the goalkeeper needs a perception (cf. Section \ref{Perception}) as his teammates in order to model the environment in which he will evolve. A concrete use of perception for cognition is the possibility to compute the intersection point between the trajectory of the ball and the future trajectory of the robot in order to intercept the ball before the goal \citep{wang2015goalkeeper}. \citet{garcia2009designing, garcia2010design} raise interesting questions about the installation of an NAO goalkeeper. \citet{garcia2009designing} explain a conflict encountered between the software performances and the hardware limits of the robot, which does not reach the intersection point with the ball trajectory in time (it takes a huge amount of time for the goalkeeper to move, almost 6 seconds per step). One year later, \citet{garcia2010design} shows good results, the ball is correctly traced, 100\% of the time, the robot is correctly positioned 84\% of the time and thus stops 62\% of the shots.

In terms of decision making, the literature contains a variety of state-machine-based approaches and learning based approaches as reinforcement learning \citep{bozinovski1999engineering}. The use of a state machine \citep{jamzad2000goal} leads to good results such as a successful shot stopping rate of 64.6\% \citep{lausen2003model}. \citet{zolanvari2019q} propose a comparison between a reinforcement learning algorithm and a hard-coded algorithm in RoboCupSoccer SSL (Small Size League), the results are in favour of learning based approaches with an average ball interception of 18.9\% for the hard-coded algorithm and 32.1\% or 53.24\% (depending on the method used) with the reinforcement learning algorithms. On the other hand a comparison between a Q-learning algorithm and a hard-coded algorithm in 3D simulation shows that learning based approach is less efficient, but could be improved \citep{fossel2010using}. The use of Q-learning coupled with a Petri Net nevertheless obtains good results \citep{gholami2017learned}. Based on \citep{lausen2003model} results the use of a state machine for decision making is a good solution. The results obtained by non-hard-coded algorithms within a simulation are also good and can be an alternative to the state machine.

Other interesting approaches are in the literature such as a Danger Theory Based strategy in 2D simulation with an efficiency of around 80\% \citep{prieto2008goalkeeper}. The two following articles follow the implementation of a goalkeeper in 1999 for the RoboCupSoccer MSL (Middle Size League) \citep{adorni1999genetic,adorni1999landmark}, the first one presents a Genetic programming approach with encouraging results on a simulator while the second one presents a landmark-based self-localisation method of the robot. Two promising approaches were published in 2015 in a single article \citep{masterjohn2015regression}. Both methods were carried out in 3D simulation. The first one proposes a deterministic behaviour using a Kalman filter based method and a linear regression. The second, based on mental simulations, predicts the future state of the environment according to its current state. The results obtained are: an average success rate of 27.23\% according to the shooting distance and 32.3\% according to the shooting angle for the deterministic algorithm and 46.8\% according to the shooting distance and 59\% according to the shooting angle for the other one.


\section{Action}
\label{Action}

\subsection{Walking}
\label{Walking}
Once the decision is made the robot must act. One of the major motion's issue for RoboCupSoccer, but also in the field of robotics in general, is navigation. How can the robot move around as efficiently as possible? We will first look at the different algorithms present in the literature setting up navigation for bipedal robots, and then we will look at the subject of stabilising the robot in order to avoid robot falling.

To enable a bipedal robot to navigate in its environment, many teams have developed algorithms using a wide range of technologies (Table \ref{table:RoboCupSoccer general developments}). One of the fastest walking speed, obtained in simulation, reaches 2.5 m/s following a reinforcement learning training \citep{abreu2019learning}. Reinforcement learning is particularly appreciated in the development of a navigation module \citep{wawrzynski2014reinforcement}. Many other algorithms were nevertheless used, publishing good results, such as a third-order interpolation offering a high stability \citep{huang1999high}, a ZMP-based method reaching a maximum speed of 10.5cm/s \citep{strom2009omnidirectional} or a Linear-Quadratic-Gaussian controller reaching a maximum speed of 80.5cm/s in simulation \citep{kasaei2019fast}. The developed walking engines are mostly omnidirectional showing an advantage within a dynamic environment \citep{behnke2006online}. These walking modules, once developed, can be optimised \citep{Reichenberg2021step}. \citet{seekircher2016adaptive} propose a LIPM-based closed-loop algorithm, used on a NAO robot, in oAdelaide, Australiarder to optimise the walk. Reinforcement learning can also be used to optimise walking \citep{tanwani2011optimizing} improving the movement speed from 1.79cm/s initially to 4.36 cm/s on a DOF bipedal robot. 

Robot falls caused by poor stabilisation can damage the robot. In order to avoid breakage and improve performance, it is necessary to overcome this problem. \citet{hohn2006detection} propose a reflex classification strategy to prevent a BARt-UH robot from falling. Combining simulation and physical robot, they obtain a stabilisation success between 22\% and 100\% when the robot is pushed with a respective force of 36Ns to 4Ns. \citet{hohn2006detection} evoke an interesting fact when talking about stabilisation: some falls are inevitable. If the robot is subjected to too strong a shock, it will be impossible for it to prevent the fall. Nevertheless, the robot's reaction to a shock can be optimised to prevent falls as much as possible. \citet{yi2011online} present a reinforcement learning algorithm on a DARwin-OP robot in order to set up a push recovery controller. Another method implemented to optimise stabilisation's controllers parameters of an IGUS humanoid robot is the Bayesian optimisation method obtaining a result of 32\% better than by-expert-tuned \citep{rodriguez2018combining}. The use of a stability controller is a major advantage in order to prevent the robot from falling.

A walking algorithm alone is not enough for a robot to evolve correctly in a dynamic environment. Path planning technologies are added to avoid dynamic obstacles such as the opposing team's robots \citep{han1997genetic}. The implementation of the walking algorithms allows the robot to dribble the ball to the goal \citep{acsik2018end} or to intercept the ball \citep{makarov2019model}.

\subsection{Kicking}
\label{Kicking}
Once the robot has a navigation module, it is very useful to have a kick module in order to pass the ball or to score goals and thus increase the chances of winning the game (cf. Table \ref{table:RoboCupSoccer general developments}).
The RoboCupSoccer is divided into several sub-leagues. Two of these sub-leagues, the SSL (Small Size League) and the MSL (Middle Size League), do not use bipedal robots and therefore have a kick system adapted to the robot hardware \citep{kasaei2010design} allowing high precision shooting \citep{houtman2019tech}. The humanoid leagues use bipedal robots and must therefore develop kick strategies taking into account the possibility of the robot falling. Coupling navigation and kick modules, \citet{pena2019adaptive} propose an omin-directional kick interpolation engine together with an adaptive walking engine used on a NAO V4 robot obtaining less than 1\% fall rate and a shooting trajectory within a six degrees cone. The deviation of the ball during the shoot is an important notion to take into account, the smaller the deviation, the more accurate the kick will be. A deflection angle of between 7.48 degrees and 8.06 degrees was obtained using Dynamic Movement Primitives (DMP) coupled with a ZMP based balancing  \citep{bockmann2016kick}. As with the navigation module, the shooting module can be managed by a controller \citep{abdolmaleki2016learning}. A robot can also learn to kick using machine learning methods. In order to allow a robot to score a goal from any part of the field in one or more shots \citep{fahami2017reinforcement} used reinforcement learning increasing significantly robot performance compared with kicking towards predetermined points in the goal. Deep Reinforcement learning was also used to learn kick behaviors improving the level of play of the robot \citep{spitznagel2021deep}.

\subsection{Goalkeeper dedicated motions}
\label{Goalkepper motions}
In order to stop the ball correctly the goalkeeper must know how to make specific movements. After determining the trajectory of the ball, the goalkeeper can perform a simple action to collide with the ball and prevent the enemy team from scoring. The goalkeeper can therefore perform a side-to-side movement, walk forward, walk backward, hold the position or dive \citep{fossel2010using}. A diving strategy for goalkeepers, published in 2018, consists in voluntarily dropping the robot so that it can stop the ball \citep{macalpine2018ut}. This strategy has been implemented in 3D simulation and is not applicable to the RoboCupSoccer SPL (Standard Platform League) without taking the risk of breaking the robot. However, it is possible for the goalkeeper to perform more complex movements, such as extending the right leg, in order to defend the largest possible area \citep{pierris2009interactive}.
 

\section{Discussion}
The first objective of this article is to bring together some of the technologies published in the RoboCupSoccer goalkeeper literature. The second objective of this article is to extract the scientific issues from this research for improving the goalkeeper behaviour. To do this we proceeded in stages. We study the different basic skills required by all players such as perception, navigation and shooting. We also looked at the developments carried out on goalkeepers. These developments are mostly oriented around decision making, which is one of the most important skills for a goalkeeper. The following analysis has emerged from this research: A simple method to implement a behavioural algorithm for a goalkeeper is the use of hard-coded algorithms such as a state machine. However, there are slightly more powerful algorithms used in simulation, but these require such computational that it is difficult to put them to real robots. The same problem is found in the field of computer vision also present in RoboCup Soccer, the algorithms were improved to reduce their computational cost \citep{ZhengBai2021faster} (section \ref{Perception}). Moreover, the results obtained in simulation are difficult to compare with the results obtained with real robots due to the difference in the environment. Nevertheless, these complex algorithms can be an advantage during a match. It is therefore conceivable to use these technologies by reducing their calculation cost. The use of a state machine remains a very good choice for the moment in terms of decision making for a non-simulation goalkeeper. Today there are several possible technology choices to implement a goalkeeper behaviour algorithm within the RoboCupSoccer domain. A simple way to create goalkeeper behaviour with correct results is to use a hard-coded algorithm, like a state machine. However, it is possible to improve this specific player by using other algorithms such as the use of linear regression coupled with a Kalman filter based method or a mental model. Unlike hard-coded algorithms, complex algorithms can predict specific unknown situations, which can be a real advantage (e.g. offside prediction with logic-based approaches). Nevertheless, there are still some blocking points for the use of these algorithms on real robots. Although these algorithms offer a good alternative to a state machine, they generally require a high computational cost to be used on a real robot. Moreover, validating these behaviours without resorting to simulation seems complex, but is still necessary to validate their good performance on real robots. The use of complex algorithms for decision making in embedded systems such as the goalkeeper remains a challenge today. The adaptation of algorithms with good results in simulation such as \citep{masterjohn2015regression} on real robots will certainly improve the performance of goalkeepers in RoboCupSoccer.


\section{Acknowledgements}
Authors would like to thank CERVVAL, Brest Métropole Océane (BMO) and AFRAN.

\bibliographystyle{unsrtnat}
\bibliography{references}  

\section{Tables}

\begin{table}[htbp]
\RaggedRight
\caption{RoboCupSoccer: Perception and Action}
\begin{tabular}{|p{2.5cm}|p{3.5cm}|p{3cm}|p{5cm}|p{1.2cm}|}
    \hline
    \rowcolor{orange!45}\textbf{Goal} & \textbf{Paper} & \textbf{Solution} & \textbf{Result} & \textbf{Robot} 
    \\\hline
    \rowcolor{blue!15}Robot and ball detection & \citep{szemenyei2019robo} & DNN & Better than Tiny YOLO in speed \& accuracy & NAOV6 
    \\\hline
    \rowcolor{blue!15}Robot and foot detection & \citep{farazi2016real} & HOG & Accuracy: Robot (88\%), Foot (89\%), Heading estimation (74\%) & IGUS 
    \\\hline
    \rowcolor{blue!25}Robot detection & \citep{cruz2017using} & CNN (XNOR-Net \& SqueezeNet) & Process in around 1 ms \& Detection rate is around 97\%  & NAO 
    \\\hline
    \rowcolor{blue!15}Robots localisation & \citep{quinlan2009multiple} & Kalman filter based & Discuss multiple-model Kalman filters and comparison with a particle filter based approach to localization & NAO 
    \\\hline
    \rowcolor{blue!25}Self localization & \citep{burchardt2010optimizing} & Particle filter based \& PSO & Results lead to a more precise position estimate & NAO 
    \\\hline
    \rowcolor{blue!15}Positioning multiple players & \citep{cano2016robust} & GM-PHD & Average error: 20cm, Processing time: 0.13ms & NAO 
    \\\hline
    \rowcolor{blue!25}3D Ball localisation & \citep{kuijpers2016cooperative} & Triangulation & Mean error: Lifted ball (0.42m), On field ball (0.11m) & MSL 
    \\\hline
    \rowcolor{blue!15}Self localisation & \citep{qian2016adaptive} & Colour segmentation \& Histogram analysis & Allows the robot to self-locate & NAOV5 
    \\\hline
    \rowcolor{yellow!25}Walk engine & \citep{strom2009omnidirectional} & ZMP-based & Max speed: 10.5cm/s & NAO 
    \\\hline
    \rowcolor{yellow!15}Closed-loop walking engine & \citep{kasaei2019fast} & Linear-Quadratic-Gaussian controller & Max speed: 80.5 cm/s & NAOsim 
    \\\hline
    \rowcolor{yellow!25}Optimise Walk & \citep{seekircher2016adaptive} & LIPM-based closed-loop & Optimized model yields a more controlled, faster and even more energy-efficient walk & NAO 
    \\\hline
    \rowcolor{yellow!15}Optimising Walking & \citep{tanwani2011optimizing} & Real-time Reinforcement learning & Optimising walking speed from 1.79cm/s to 4.36cm/s & DOF 
    \\\hline
    \rowcolor{yellow!25}Classify \& Prevent falls & \citep{hohn2006detection} & Reflex classification & Stabilisation success between 22\% and 100\% from impulse between 36Ns and 4Ns & BARt-UH 
    \\\hline
    \rowcolor{yellow!15}Push recovery controller & \citep{yi2011online} & Reinforcement learning & This online method can stabilize an inexpensive, commercially- available DARwin-OP small humanoid robot & DARwin-OP 
    \\\hline
    \rowcolor{yellow!25}Optimise stabilisation's controllers parameters & \citep{rodriguez2018combining} & Bayesian optimisation method & 32\% better than by-expert-tuned & IGUS 
    \\\hline
    \rowcolor{green!15}Walk-kick framework & \citep{pena2019adaptive} & Kick interpolators \& Adaptive walking engine&Less than 1\% falling rate \& Walk-kick trajectories less than 6 degree within any given direction & NAOV4 
    \\\hline
    \rowcolor{green!25}Kick motions & \citep{bockmann2016kick} & Dynamic Movement Primitives (DMP) \& ZMP based balancing & Work but less far shoot, Angle deviation between 7.48deg and 8.06deg & NAO 
    \\\hline
    \rowcolor{green!15}Kick controller & \citep{abdolmaleki2016learning} & Contextual policy search method CREPS-CMA & Average error from linear (0.82 +/- 0.10m) to non-linear (0.34 +/- 0.11m) policy & NAOsim 
    \\\hline
    \rowcolor{green!15}Learn  kick  behaviors  & \citep{spitznagel2021deep} & Deep Reinforcement learning & A significant improvement of  the general level of play & NAOsim 
    \\\hline
\end{tabular}
\label{table:RoboCupSoccer general developments}
\end{table}

\begin{table}[htbp]
\RaggedRight
\caption{RoboCupSoccer: Goalkeeper Cognition}
\begin{tabular}{|p{2.5cm}|p{3.5cm}|p{3cm}|p{5cm}|p{1.2cm}|}
    \hline
    \rowcolor{orange!45}\textbf{Goal} & \textbf{Paper} & \textbf{Solution} & \textbf{Result} & \textbf{Robot} 
    \\\hline
    \rowcolor{cyan!15}Goalkeeper for MSL & \citep{jamzad2000goal} & State machine & A goalkeeper misses the ball,basically in two situations: Not seeing the ball and seeing the ball but not being able to move fast enough to catch it & MSL 
    \\\hline
    \rowcolor{cyan!25}Design a MSL Goalkeeper & \citep{lausen2003model} & State machine & 64.6\% of goals saved& MSL 
    \\\hline
    \rowcolor{cyan!15}Evolving goalkeeper behaviour & \citep{bozinovski1999engineering} & Hand-coded \& Q-learning & Performance ratio between 4 and 10 (ratio 3 = 75\% success rate) & 
    \\\hline
    \rowcolor{cyan!25}Penalty kick goalkeeping for SSL & \citep{zolanvari2019q} & Hard coded vs Q-learning & Average Success rate (simulation): Hard coded (18.9\%), NR(32.1\%), EBR(53.24\%) & SSL 
    \\\hline
    \rowcolor{cyan!15}Build a goalie behaviour & \citep{prieto2008goalkeeper} & Danger Theory Based strategy & Effectiveness above 80\% & 2Dsim 
    \\\hline
    \rowcolor{cyan!25}Motion-control strategy for a goalkeeper MSL & \citep{adorni1999genetic} & Genetic programming (GP) & The preliminary results obtained on a simulator are encouraging & MSL 
    \\\hline
    \rowcolor{cyan!15}Design Goalkeeper & \citep{adorni1999landmark} & Landmark-based robot self-localisation & The robot is able to perform its tasks in real time & MSL 
    \\\hline
    \rowcolor{cyan!25}MSL Goalkeeper, Detect positions of the ball to predict the interception point & \citep{wang2015goalkeeper} & Image processing \& Hand-coded algorithm & The experimental results show the accuracy and effectiveness of this strategy & MSL 
    \\\hline
    \rowcolor{red!15}Goalkeeper behaviour & \citep{garcia2009designing} & State machine & Conflict between software performance and hardware capacity, the goalkeeper can't reach the point in time & NAO 
    \\\hline
    \rowcolor{red!25}Goalkeeper behaviour & \citep{garcia2010design} & State machine & The architecture allowed the robot to track the ball nearly 100\% of the time, position itself properly 84\% of the time, and save goals from 62\% of the trajectories tested & NAO 
    \\\hline
    \rowcolor{red!15}Goalkeeper strategy & \citep{fossel2010using} & Q-learning vs Hard coded & The Q-learning algorithm is less efficient but can be enhance & NAOsim 
    \\\hline
    \rowcolor{red!25}New learning-based behaviour model for a soccer goalkeeper & \citep{gholami2017learned} & Petri nets \& Q-learning & Results of theoretical analysis of some case studies show impressive performance improvement in goalkeeper task execution & 
    \\\hline
    \rowcolor{red!15}Goalkeeper strategy & \citep{macalpine2018ut} & Diving behaviour & Success rate 46.6\% & NAOsim 
    \\\hline
    \rowcolor{red!25}Goalkeeper strategy sim 3D & \citep{masterjohn2015regression} & Deterministic behaviour (Linear Regression/Kalman filter) \& Mental model approach & Success rate: Deterministic behaviour: distance ~27,23\%, angle ~32,3\%, mental model approach: distance ~46,8\%, angle ~59\%. & NAOsim 
    \\\hline
\end{tabular}
\label{table:RoboCupSoccer Goalkeeper developments}
\end{table}

\end{document}